\documentclass[runningheads]{llncs}
\usepackage{amsmath} 
\usepackage{amssymb}
\usepackage{boondox-cal} 
\usepackage{mathrsfs}
\usepackage{hyperref}
\hypersetup{colorlinks, 
            citecolor=blue,
            filecolor=black,
            linkcolor=black,
            linktocpage,
            urlcolor=black}
\usepackage{graphicx}
\usepackage{tikz}
\usepackage{subcaption}
\usepackage{bayesnet}
\usepackage[misc]{ifsym}
\usepackage[separate-uncertainty=true]{siunitx}
\usepackage{etoolbox,booktabs}

\newcommand{\vect}[1]{\boldsymbol{#1}}

\begin{document}
\title{Conditional Neural Relational Inference for Interacting Systems}
%
\author{
  Joao A. Candido Ramos\inst{1}\inst{2}\Letter
  \and Lionel Blondé\inst{1}\inst{2}
  \and Stéphane Armand\inst{1}
  \and Alexandros Kalousis\inst{2}
}
\authorrunning{
  Joao A. Candido Ramos et al.
}
%
\institute{University of Geneva, Switzerland \and Geneva School of Business Administration, HES-SO, Switzerland\\
\email{joao.candido@etu.unige.ch}\\} %

\toctitle{Conditional Neural Relational Inference for Interacting Systems}
\tocauthor{J. Candido Ramos et al.}
\maketitle		
%

%
%
%
%

\begin{abstract}
  In this work, we want to learn to model the dynamics of similar yet distinct groups of interacting objects.
  These groups follow some common physical laws that exhibit specificities that are captured through some vectorial description.
  We develop a model that allows us to do conditional generation from any such group given its vectorial description.
  Unlike previous work on learning dynamical systems that can only do trajectory completion and require a part of the trajectory dynamics to be provided as input in generation time, we do generation using only the conditioning vector with no access to generation time's trajectories.
  We evaluate our model in the setting of modeling human gait and, in particular pathological human gait.
\end{abstract}

\section{Introduction}

While modeling the evolution of an object in a physical dynamical system
already constitutes a tedious endeavor, modeling the evolution of a system of
objects interacting with each other is considerably more challenging.
The complex physical laws describing the system are, in most cases, unknown to
the learning agent, who then only has access to observations depicting traces
of interaction of the whole physical system, called trajectories.
Previous works have attempted to learn the dynamics of systems involving
interacting objects by injecting a structural inductive bias in the model,
allowing them to learn the inter-object relationships~\cite{Battaglia2016,Chang2016,Hoshen2017,Santoro2017,VanSteenkiste2018,Watters2017,Kipf2018,Webb2019}.
When the relationships between the interacting objects are unknown 
\textit{a priori}, there exist two approaches to leverage the lack of structural
information: modeling the interactions implicitly or explicitly.
The first approach describes the physical system by a fully connected graph
where the message passing operations implicitly describe
the interactions, hoping that useful connections will carry more information~\cite{Guttenberg2016,Santoro2017,Sukhbaatar2016,Watters2017}.
Other works add an attention mechanism to give more importance to some
interactions in the fully connected graph~\cite{Hoshen2017,VanSteenkiste2018}.
In the second approach, we have unsupervised models, such as NRI~\cite{Kipf2018} and fNRI~\cite{Webb2019},
which can explicitly predict the interactions and dynamics of a physical system
of interacting objects only from their observed trajectories.
When it comes to predicting the future states of the system, previous works adopt different strategies.

In the prediction of the future states of the physical system, we find different strategies.
Some works predict the next state from the previous ones~\cite{Battaglia2016,Chang2016}.
Others, such as NRI, predict the continuation of the trajectories given a first observed part of the trajectories, essentially doing trajectory completion.
All of them require access to a part of the trajectories to make the prediction of the next states~\cite{Kipf2018,Webb2019}. 
To the best of our knowledge, there is no work that considers how the specificities of a given physical system impact the dynamics learned by such models, 
as well as how expliciting them through a conditioning feature vector can result in generated trajectories displaying the specific fingerprint behavior of the considered examples.
In this work, we want to solve the problem of learning several slightly different dynamical systems, where the information differentiating them is contained in a description vector.
To illustrate our setting, let us consider the modelling of human gait which has driven this work.
Human gait follows a certain number of biomechanical rules that can be described in terms of kinetics and kinematics but also depends to a considerable extent on the individual.
The neurological system and the person's past may influence the manner the individual walks significantly.

To generate trajectories for a given group of interacting objects, we introduce a conditional extension of NRI (cNRI) that can generate trajectories from an interaction graph given a conditioning vector describing that group.
By providing the conditioning vector to the decoder, we allow the encoder to be any model that can output interactions.
The decoder learns to generate the dynamics of the physical system from the conditioning vector.
The encoder can be a fixed, known, graph, i.e. it is not learned, similar to the true graph in the original paper~\cite{Kipf2018}.
Our work differs considerably from NRI; we do not seek to learn the interactions explicitly.
Instead, we want to use these interactions, whether they are given or learned, together with the conditioning vector 
to conditionally generate trajectories given only the conditioning vector.

We demonstrate our approach in the problem of learning to conditionally generate the gait of individuals with impairments.
The conditioning vector describes the properties of an individual. Our ultimate goal is to provide decision support for 
selecting the appropriate treatment (surgery) for any given patient; this work is a stepping stone towards that direction. 
Selecting the most appropriate surgery for patients with motor and neurological problems such as cerebral palsy is a challenging 
task~\cite{Pitto2019}.  A tool that can model pathological gait and conditionally generate trajectories can allow physicians to 
simulate the outcomes of different operations on the patient's gait simply by modifying the conditioning vector. 
This will reduce in a considerable manner unnecessary operations and operations with adverse effects.

We will learn the dynamics of gait from the set of trajectories of the different body parties, described 
either in the form of euclidean coordinates or as joint angles.  The conditioning vector will contain 
clinical information describing the patient's pathology, their anthropometric parameters, and measurements 
acquired during a physical screening. We experimentally show that our model achieves the best results in 
this setting, outperforming in a significant and consistent manner relevant baselines, providing thus a 
promising avenue for eventual decision support for treatment selection in the clinical setting. 

\section{Related Work}

There are many works that tackle the problem of motion forecasting, using traditional methods such as hidden Markov models~\cite{Lehrmann2014}, 
Gaussian process latent variable models~\cite{Wang2006,Urtasun2008} or linear dynamical systems~\cite{Pavlovic2001}.
More recently, recurrent networks have been used to predict the future positions in a sequential manner~\cite{Fragkiadaki2015,Jain2016,Martinez2017,Walker2017,Gopalakrishnan2018,Liu2019,Li2019,Aliakbarian2020ASC}.
Imitation learning algorithms have also been used to model human motion~\cite{Wang2019}.
However, all previous attempts use a part of the trajectories to predict their future. 
To the best of our knowledge, no work tackles the problem of full trajectory generation 
conditioned only on a description of the system for which we wish to generate trajectories.

\section{The Conditional Neural Inference Model}\label{sec:model}

We want to learn to model the dynamics of multi-body systems consisting of $M$ interdepedent and interacting bodies. Such a system when it evolves
in time it generates a multi-dimensional trajectory $\mathbf{X} = [\mathbf{x}^1, ..., \mathbf{x}^T]$ (we assume trajectories of fixed length $T$),
where the $\mathbf x^t$ element of that trajectory is given by $\mathbf{x}^t = [\mathbf{x}^t_1, ..., \mathbf{x}^t_M]^\text{T}$ 
and $\mathbf x^t_i$ is the set of features describing the properties of the $i$ body at time $t$. We will denote the complete
trajectory of the body-part $i$ by $\mathbf x_i^{1:T}$.
In the following we will use
boldface to indicate samples of a random variable and caligraphic for the random variable itself.
One example of such an  
$\mathbf x^t_i$ can be the euclidean coordinates of the $i$th body if the trajectories track position of the body parts of a 
multi-body system. In addition each such system is also described by a set of properties $\mathbf c \in \mathbb R^d$ providing
high level properties of the system that determine how its dynamics will evolve.
Our goal is to learn the conditional generative model $p(\mathcal X | \mathcal c)$ which will allow us to generate 
trajectories given only their conditioning property vector $\mathbf c$. Our training data consist of pairs $(\mathbf X_i, \mathbf c_i), i:=1 \dots N$,
produced by $N$ different dynamical systems. Since we base our model on the NRI we will first provide a brief description of it.

In NRI the goal is to learn the dynamics of a {\em single} multi-body dynamical system and use the learned dynamics to forecast 
the future behavior of trajectories sampled from that system. To solve the forecasting problem it learns a latent-variable generative model of $P(\mathcal X)$ where the latent variable captures the 
interactions.  The training data $\mathbf X_i, i:=1, \dots, N,$ are thus samples from a fixed dynamical system whose dynamics NRI will learn. The basic NRI model is a Variational Auto-Encoder (VAE), \cite{Kingma2014}.
The latent representation is a matrix-structured latent variable $\mathcal Z : N \times N$, where ${\mathcal z}_{i,j}$ is a $K$-category categorical random variable describing the type of interaction, 
if one exists, between the $i,j$, bodies of the system. The approximate posterior distribution is given by  $q_\phi(\mathcal Z | \mathcal X) = \prod_{i,j} q_\phi( \mathcal{z}_{i,j} | \mathcal X)$, 
where $ \mathcal{z}_{i,j} \sim q_\phi( \mathcal{z}_{i,j} | \mathcal X) = \text{Cat}(\mathbf p=[p_1,...,p_K]=\vect \pi_{\phi_{i,j}}(\mathcal X))$. 
The encoder $\vect \pi_{\phi}(\mathcal X)$ is a graph network that feeds on the trajectory data 
and outputs the probability vector for each $i,j$, interaction based on the learned representation of the respective $i,j$, 
edge of the graph network; more details on the encoder in Section \ref{par:nri_encoder}.

The generative model has an autoregressive structure given by: 
$p_{\theta} (\mathcal X | \mathcal Z) = \prod_{t=1}^{T} p_{\theta} ( \mathcal{x}^{t+1} | \mathcal { x}^{t}, ..., \mathcal {x}^{1}, \mathcal Z )$, 
where $p_{\theta} ( \mathcal {x}^{t+1} | \mathcal {x}^{t}, ..., \mathcal {x}^{1}, \mathcal Z )$  
$ = \mathcal N (\vect \mu_{\theta}(\mathcal {x}^{t}, ..., \mathcal {x}^{1}, \mathcal Z)$, $\sigma^2 \mathbf I) $.
The $\mu_{\theta}(\mathcal {x}^{t}, ..., \mathcal {x}^{1}, \mathcal Z)$ is a graph network that feeds
on the learned interaction matrix and the so far generated trajectory\footnote{The first part of that 
trajectory will always be real data, even at test time, directly coming from the input trajectory 
as we will soon explain.}. The autoregressive model in the generative 
distribution is trained using teacher forcing up to some step $l$ in the trajectory after which the predictions are used to generate 
the remaining trajectory points from $l+1$ to $T$. This is a rather important detail because it also reflects how the decoder is used at test time to do trajectory forecasting.
At test time in order for NRI to forecast the future of a given trajectory it will feed on the trajectory and then map it to its latent representation.
Its decoder will feed on the real input trajectory and thanks to its autoregressive nature will generate its future states.
By its conception NRI does not learn over different dynamical systems, nor can it generate trajectories from scratch, it has to feed on trajectory parts and then forecast.  To 
address this setting we develop a conditional version of NRI.

The conditional-NRI (cNRI) has the same model architecture as NRI, i.e. it is a VAE with an encoder that outputs a latent space, structured as above, that describes the interactions and a decoder 
generates the complete trajectories. Unlike NRI which learns the distribution $p(\mathcal X)$ of a fixed dynamical system here we want to learn over different dynamical systems and be able to generate 
from trajectories at will from each one of them. Thus in cNRI we model the conditional distribution $p(\mathcal X| \mathcal c)$ where $\mathcal c$ provides the description of the conditional generation 
system from which we wish to sample. The posterior distribution is the same as that of NRI, while the generative distribution is now  $p_{\theta} (\mathcal X | \mathcal Z, \mathcal c)
= \prod_{t=1}^{T} p_{\theta} ( \mathcal {x}^{t+1} | \mathcal {x}^{t}, ..., \mathcal {x}^{1}, \mathcal {Z} , \mathcal {c})$, where $p_{\theta} ( \mathcal {x}^{t+1} | \mathcal {x}^{t}, ..., \mathcal {x}^{1}, 
\mathbf Z , \mathcal c) = \mathcal N (\vect \mu_{\theta}(\mathcal {x}^{t}, ..., \mathcal {x}^{1}, \mathcal Z, \mathcal c), \sigma^2 \mathbf I) $. 
Unlike NRI we train the decoder without teacher forcing; at test time when we should conditionally generate a trajectory $\mathbf X$ from the 
description $\mathbf c$ of a dynamical system we do not require access to any trajectory from that system. 

Our loss is the standard ELBO loss adjusted for the conditional setting and the optimization problem is:
\begin{equation}
	\label{optimization.prob}
	\max_{\phi, \theta} \mathbb E_{\mathbf X, \mathbf c ~\sim P(\mathcal X, \mathcal c)}
        \mathbb{E}_{\mathbf Z \sim q_\phi({\mathcal Z}|{\mathbf{X}})} [\log p_\theta({\mathbf X}|{\mathbf Z}, {\mathbf c})] - D_{\text{KL}}[q_\phi({\mathbf Z}|{\mathbf{X}})||p({\mathbf Z})]
\end{equation}

\begin{figure}
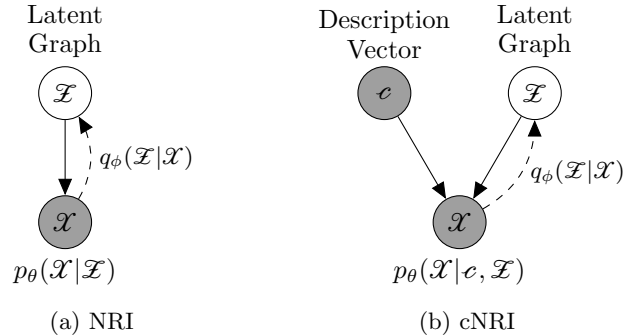

  \centering
  \begin{subfigure}[b]{0.4\textwidth}
    \centering
    \tikz{
      \node[obs, label=below: {\normalsize \(p_\theta(\mathcal{ X } | \mathcal Z)\) } ] (x) {$\mathcal X$};%
      \node[latent, above=of x, label={\normalsize Latent Graph}] (z) {$\mathcal Z$}; %
      \edge {z} {x} ;
      \path (x) edge [->, >={triangle 45}, dashed, bend right=30] node[right]{\(q_\phi(\mathcal Z | \mathcal X)\)} (z);
    }
    \caption{NRI}
  \end{subfigure}
  \begin{subfigure}[b]{0.4\textwidth}
    \centering
    \tikz{
      \node[obs, label=below: {\normalsize \(p_\theta(\mathcal X| \mathcal {c}, \mathcal Z)\)}] (x) {$\mathcal X$};%
      \node[obs, above=of x, xshift=-1cm, label={\normalsize Description Vector}] (c) {$\mathcal c$}; %
      \node[latent,above=of x,xshift=1cm, label={\normalsize Latent Graph}] (z) {$\mathcal Z$}; %
      \edge {c, z} {x} ;
      \path (x) edge [->, >={triangle 45}, dashed, bend right=30] node[right]{\(q_\phi( \mathcal Z | \mathcal X)\)} (z);
    }
    \caption{cNRI}
  \end{subfigure}
  \caption{The NRI and cNRI graphical models.}
	\label{fig:graph_model}%
\end{figure}
In the following sections we will review different options for the encoder architecture and we will described the decoder's architecture. 

\subsection{Encoding, establishing the body-part interactions}
In NRI the role of the encoder is to learn the interaction network which is then used in the decoder to support the learning of the dynamics. 
However, in cNRI the primary goal is not to learn the interaction graph but to be able to conditionally generate trajectories from different 
dynamical systems. We will thus explore and evaluate different scenarios with respect to the prior knowledge we have about the interaction 
graph. In particular we will consider scenarios in which the real interaction graph is known and scenarios in which it is unknown and we learn it.  
Strictly speaking in the former case we do not have an encoder anymore and we are not learning a variational autoencoder but rather a conditional
generative model that explicitly maximizes the data likelihood.

\paragraph{Perfect interaction graph}
In this scenario we assume that the interaction graph $\mathbf Z$ is known and it is the same for all our different dynamical systems.
So in that setting there is no encoder involved, or alternatively we can think of the encoder as a constant function that maps all 
instances to the same latent vector. As an example in the gait modelling problem the $\mathbf Z$ matrix will be the adjacency matrix 
that describes the body-parts connectivities as these are given by the human skeleton. 
So in this setting the optimization problem reduces to:
\begin{equation}
	\label{optimization.prob.perfect}
	\max_{\theta} \mathbb E_{\mathbf X, \mathbf c ~\sim P(\mathcal X, \mathcal c)} [\log p_\theta({\mathbf X}|{\mathbf Z}, {\mathbf c})]                                                                  
\end{equation}

\paragraph{Imperfect interaction graph}
There are cases in which we have a good understanding of the interaction between the different body-parts but we do not have 
the complete picture. If we turn back to the example of the human gait modelling, the interactions between the body parts are 
not only sort range, through the immediate connections as above, but also longer range; while walking our arms in the opposite 
directions as the feet of the opposite body side.  When we model the dynamics on the decoder side it might be beneficial for 
the generations to have explicitly in the interaction graph such longer dependencies. Remember that the decoder is a graph network
whose adjacency matrix is given by $\mathbf Z$, having the longer dependencies explicitly modelled will not require the decoder 
graph network to transfer information over longer paths. To account for such a setting we now make $\mathbf Z$ a learnable
parameter starting from the original interaction graph. As before there is no encoder and learning $\mathbf Z$ consists in 
making the generative model a function of the $\mathbf Z$ which is not sampled from the posterior distribution but treated
as a deterministic variable that we learn with standard gradient descent. So our generative distribution is now
 $p_{\theta} (\mathcal X | \mathbf Z)$. Such an approach has been also used in~\cite{Shi2019}. The optimization 
problem now is:
\begin{equation}
	\label{optimization.prob.imperfect}
	\max_{\mathbf Z, \theta} \mathbb E_{\mathbf X, \mathbf c ~\sim P(\mathcal X, \mathcal c)}
                                                                      [\log p_\theta({\mathbf X}|{\mathbf Z}, {\mathbf c})]                                                                  
\end{equation}

\paragraph{Unknown interaction graph, the NRI encoder.}\label{par:nri_encoder}
Often the interaction graph is not known. NRI was proposed for exactly such settings. Its 
encoder, $\vect \pi_\phi (\mathbf X)$, a fully connected graph network, learns the parameters 
of the categorical posterior distribution from which the latent interaction graph is sampled 
from the complete trajectories of the different body parts.  
In particular  $\vect \pi_\phi (\mathbf X)$ consists of the following message passing operations:
\begin{align*}
  \small
  {\mathbf h}_j^0   &= f_{emb}({\mathbf x}_j^{1:T}),                    & {\mathbf h}^1_{(i,j)}  &= f_e^1([{\mathbf h}_i^0, {\mathbf h}_j^0]),       \\
  {\mathbf h}^{1}_j &= f_v^1([ \sum_{i \neq j} {\mathbf h}^1_{(i,j)}]), &  {\mathbf h}^2_{(i,j)} &= f_e^2([{\mathbf h}_i^1, {\mathbf h}_j^1]), & \vect \pi_{\phi_{i,j}} (\mathbf X) & = \text{Softmax}({\mathbf h}^2_{(i,j)})
\end{align*}
$f_{emb}({\mathbf x}_j^{1:T})$ is a neural network that learns a hidden representation of the body-part (node) $j$ from its 
full trajectory; $f_e^1([{\mathbf h}_i^0, {\mathbf h}_j^0])$ is a network that learns a hidden representation 
of the edge connecting nodes $i$ and $j$;  $f_v^1([ \sum_{i \neq j} {\mathbf h}^1_{(i,j)}])$ updates the representation of the $j$ 
node using information from all the edges in which it participates and finally $f_e^2([{\mathbf h}_i^1, {\mathbf h}_j^1])$ is a
network that computes the final $K$-dimensional edge representation. This final representation of the $i,j$ edge is passed from 
a softmax function to give the proportions $\mathbf p$ of the categorical distribution $q_\phi(\mathcal z_{i,j} | \mathcal X)$ 
from which we sample the type of the i,j edge. 

With this formulation, the encoder has to assign an edge-type per pair of nodes, 
preventing the model from generalizing well on problems where the interaction graph should be sparse. As a solution~\cite{Kipf2018} 
proposes defining an edge type as a non-edge, so no messages are passing through it.

\paragraph{Unknown interaction graph, the fNRI Encoder}
In certain cases one might want more than a single edge type connecting at the same time a given pair of nodes. In the standard NRI approach this is not 
possible since the edge type is sampled from a categorical distribution. Instead we can model the $\mathcal {z}_{i,j}$ variable as a $K$-dimensional random 
variable whose posterior  $q_\phi(\mathcal z_{i,j} | \mathcal X)$ is given by a product of $K$ Bernoulli distributions and have the graph network learn the 
parameters of these $K$ distributions. More formally:
\begin{align*}
q_\phi(\mathcal z_{i,j,k} | \mathcal X) = \text{Ber}(p_{k_{i,j}}=\vect \pi_{\phi_{i,j,k}}(\mathcal X))
\end{align*}  
This is the approach taken in factorised NRI (fNRI) proposed in~\cite{Webb2019}. 
Instead of passing the result of the first message passing operation \({\mathbf h}^{1}_j\) 
through the second edge update function as NRI does, fNRI uses \(K\) edge update functions 
to get \(K\) different two-dimensional edge embeddings \(h^2_{(i,j)}\) which are passed from 
a $\text{Softmax}$ function to get the parameters of the $K$ Bernoulli distributions:

\begin{align*}
  {\mathbf h}^{2,l}_{(i,j)} & =f_e^{2,l}([{\mathbf h}_i^1, {\mathbf h}_j^1]), & {\mathbf h}^{2}_{(i,j)} & = [{\mathbf h}^{2,1}_{(i,j)}, ..., {\mathbf h}^{2,K}_{(i,j)}], & 
  \pi_{\phi_{i,j,k}}(\mathbf  X)  & = \text{Softmax}({\mathbf h}^{2,k}_{(i,j)}) 
\end{align*}

When we learn the interaction graph using the NRI or the fNRI encoders we are sampling from a categorical 
distribution. In order to be able to backpropagate through the discrete latent variable $\mathbf Z$
we use their continuous relaxations given by the concrete distribution~\cite{Maddison2016}:
\begin{equation}
	z_{i,j} = \text{Softmax}(\frac{h^2_{(i,j)} + \mathbf{g}}{\rho})   \ \ \ \ 
	z_{i,j,k} = \text{Softmax}(\frac{h^{2,k}_{(i,j)} + \mathbf{g}}{\rho})  
\end{equation}
where \(\mathbf{g}\) is a vector of i.i.d samples from the Gumbel(0,1) distribution and 
\(\rho\) is the temperature term.

\subsection{Decoding, establishing the dynamics}
The role of the decoder is to learn the dynamics so that it can successfully generate trajectories for any given dynamical system.
As already discussed the NRI architecture is designed for forecasting and does not address this task. This is because at test time 
in order to establish the interaction matrix its encoder needs to feed on a trajectory of the given system and the decoder needs this trajectory 
in order to achieve the forecasting task. In our setting at test time we do not have access to the trajectories but only to the condition 
vectors $\mathbf c$ of some dynamical system. The generative model of cNRI will only feed on the conditioning vector, the interaction 
matrix, and the initial state $\mathcal x^1$ that provides a placement for the trajectory, and it will unroll its autoregressive structure
only over generated data, more formally:
$p_{\theta} (\mathcal X | \mathcal Z, \mathcal c, \mathcal x^1) = \prod_{t=1}^{T} p_{\theta} ( \mathcal{x}^{t+1} | \hat{\mathcal {x}}^{t}, ..., \hat{\mathcal {x} }^{2}, \mathcal {x}^{1}, \mathcal c, \mathcal Z )$, 
where $\hat{\mathcal x}^{t}$ is the $t$ state of the trajectory sampled from the generative model.

To condition the generative model on the conditioning vector $\mathbf c$ we bring the information of the
conditioning vector in two places within the generative model. First when to learn the initial hidden states
of the different nodes (body-parts) we use an MLP that feeds on $\mathbf c$ and outputs an embedding 
 \({\mathbf h}^{0} = f_c^{\text{hid}}(\mathbf c)\) of size \(N \times H\) where \(H\) is the number of hidden 
dimensions we use to represent each one of the $N$ nodes; as a result each $i$ node has its 
own representation ${\mathbf h}^{0}_i$ which does not require the use of trajectory information. 
In NRI the node embeddings are initialized with zero vectors and the input trajectory is used as 
burn-in steps to update the state embeddings before forecasting the future trajectory.

One problem with the above conditioning is that it is used to compute only the initial hidden state of each node, whose effect due to the autoregressive
nature of the decoder can be eventually forgotten. To avoid that we also use the conditioning vector \(\mathbf c\) directly inside the message passing 
mechanism of the decoder. To do so we create a virtual edge that is a function of the conditioning vector and links to every node; essentially
the conditioning vector becomes a global attribute of the graph that is then used by all update functions~\cite{Battaglia2018}.
The virtual edge embedding is computed through an MLP as \({\mathbf h}^{\text{msgs}}=f_c^{\text{msgs}}\) and used in updating the stats of all nodes.

When we use the fNRI encoder the decoder $\mu_{\theta}(\hat{\mathbf {x}}^{t}, ..., \hat{\mathbf {x} }^{2}, \mathbf {x}^{1}, \mathbf c, \mathbf Z)$ 
performs the following messages-passing and autoregressive operations to get the mean of the normal distribution from which the next trajectory state is sampled:
\begin{align*}
  {\mathbf h}^{t}_{(i,j)} & = \sum_{k} z_{ij,k} f_e^k([{\mathbf h}_i^t, {\mathbf h}_j^t]) &
  \bar{\mathbf h}_j^t & = {\mathbf h}^{\text{msgs}} + \sum_{i \neq j} {\mathbf h}^t_{(i,j)}\\
  {\mathbf h}^{t+1}_{j} & = \text{GRU}([\bar{\mathbf h}_j^t, \hat { {\mathbf x}}^t_j], {\mathbf h}_j^t) & 
  {\vect \mu}^{t+1}_j        & = {\mathbf x}^t_j + f_{out}({\mathbf h}^{t+1}_{j})
\end{align*}
where ${\mathbf h}^{t}_{(i,j)}$ is the hidden representation of the $i,j$, edge at time $t$ computed from the 
hidden representations of the $i$, $j$, nodes it connects. Note that this takes into account all different edge
types that connect $i$ and $j$ through the use of one edge update function $f_e^k$ per edge type. The $z_{ij,k}$ 
variable acts as a mask. If we use the NRI encoder then only one edge update is selected, since in that case 
there can be only one edge type connecting two nodes. 
$\bar{\mathbf h}_j^t$ is the aggregated edge information that arrives at node $j$ at time $t$ computed from all edges 
that link to it as well as the virtual edge. The new hidden state of the node $j$, ${\mathbf h}^{t+1}_{j}$, is
given by a GRU which acts on the sampled $\hat { {\mathbf x}}^t_j$, the respective hidden representation ${\mathbf h}_j^t$, 
and the aggregated edge information,  $\bar{\mathbf h}_j^t$. From this ${\mathbf h}^{t+1}_{j}$ we finally
get the mean of the normal distribution from which we sample the next state of the trajectory as shown above; essentially
we use the hidden represtation to compute an offset from the previous state through the $f_{out}$ MLP. 

\subsection{Conditional Generation}
Once the model is trained we want to use the generative model $p_{\theta} (\mathcal X | \mathcal Z, \mathcal c, \mathcal x^1)$ to conditional generate 
trajectories from a dynamical system for which we only have access to $\mathbf c$ but not its trajectory, in such a case the interaction graph is not known.
We sample the $\mathbf Z$ from the aggregate posterior : $q_\phi(z) = q^{\text{avg}}_\phi(z) \triangleq \frac{1}{N}\sum^N_{n=1} q_\phi(z|x_n)$.
Since we have a discrete distribution, the aggregated posterior is the probability to have a given edge-type in training samples.
The sampling of the interaction graph only occurs in the unsupervised encoders.
Finally to simplify our evaluation, we are not learning the probability of \(p(\mathcal x^1|\mathcal c)\). 
We are giving this frame as the starting point of the generations.
Nevertheless this probability can be learned by a neural network or by the decoder directly.

\section{Experiments}\label{sec:experiments}
As we have discussed in the introduction the main motivation for this work is the provision of decision support for the treatment of 
patients with motor impairements where the conditioning vector describes how an operation affects body structure and the generative
model will show how such changes affect gait. 

The data have been collected from a kinesiology laboratory and come from patients with hemiplegia. They contain
the kinematics and clinical data of 72 patients for a total of 132 visits at the laboratory. The kinematics data
are recorded by placing markers on the body patient who then walks on a corridor where infrared cameras record the
motion. The clinical data, our conditioning vector $\mathbf c$, are obtained by a physiotherapist and include 
parameters such as body measurements and evaluation of muscles' strength; we have a total of 84 such parameters. 
From the available data we obtain 714 gait cycles, where each cycle is a multidimensional trajectory giving 
the position of all body parts through time. 

\begin{figure}[ht]
\begin{center}
  \begin{subfigure}{.2\textwidth}
    \centering
    \includegraphics[width=\linewidth]{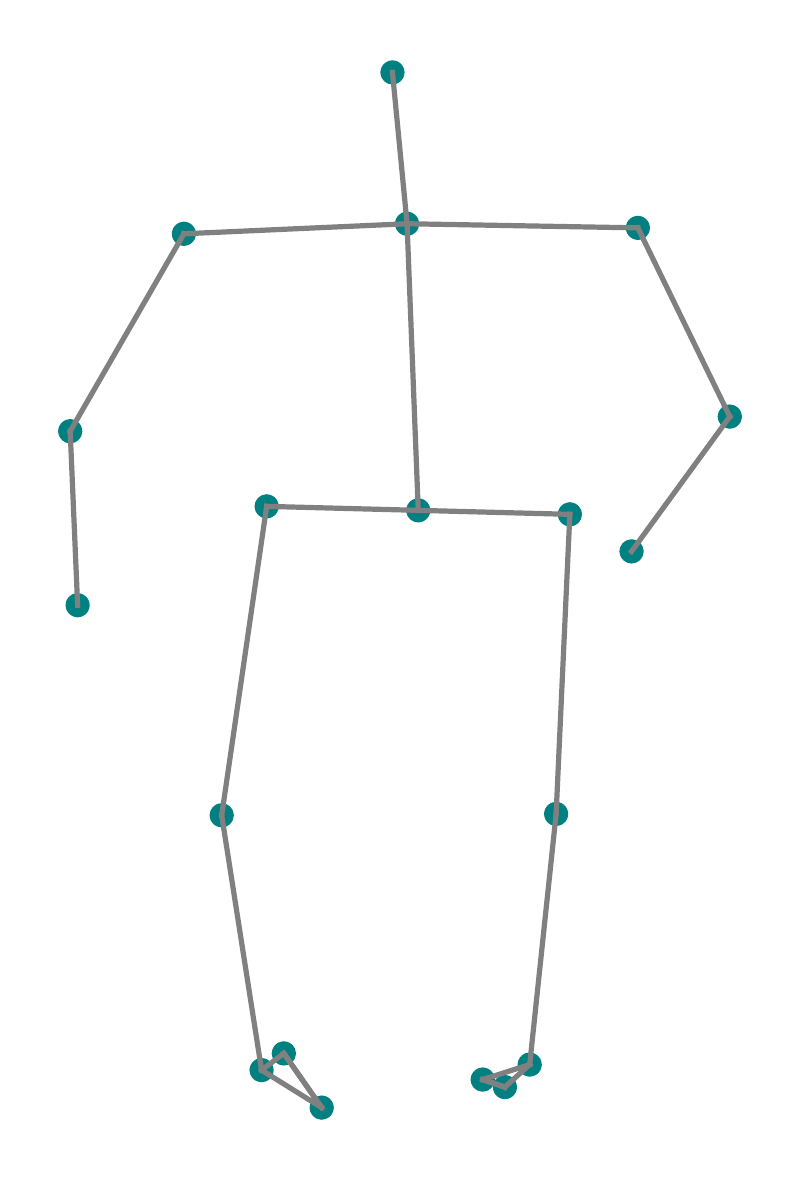}
    \caption{Skeleton}\label{fig:skeleton}
  \end{subfigure}
  \begin{subfigure}{.2\textwidth}
    \centering
    \includegraphics[width=\linewidth]{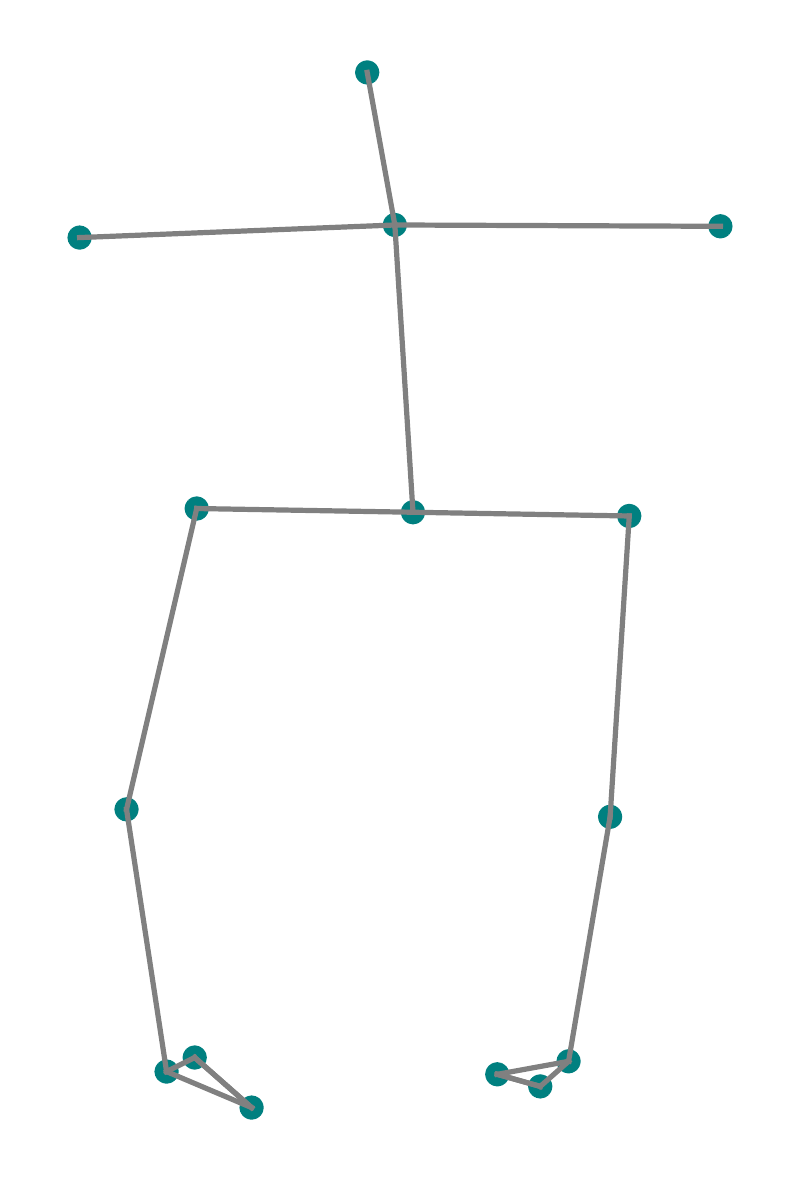}
    \caption{Armless}\label{fig:armless}
  \end{subfigure}
  \begin{subfigure}{.2\textwidth}
    \centering
    \includegraphics[width=\linewidth]{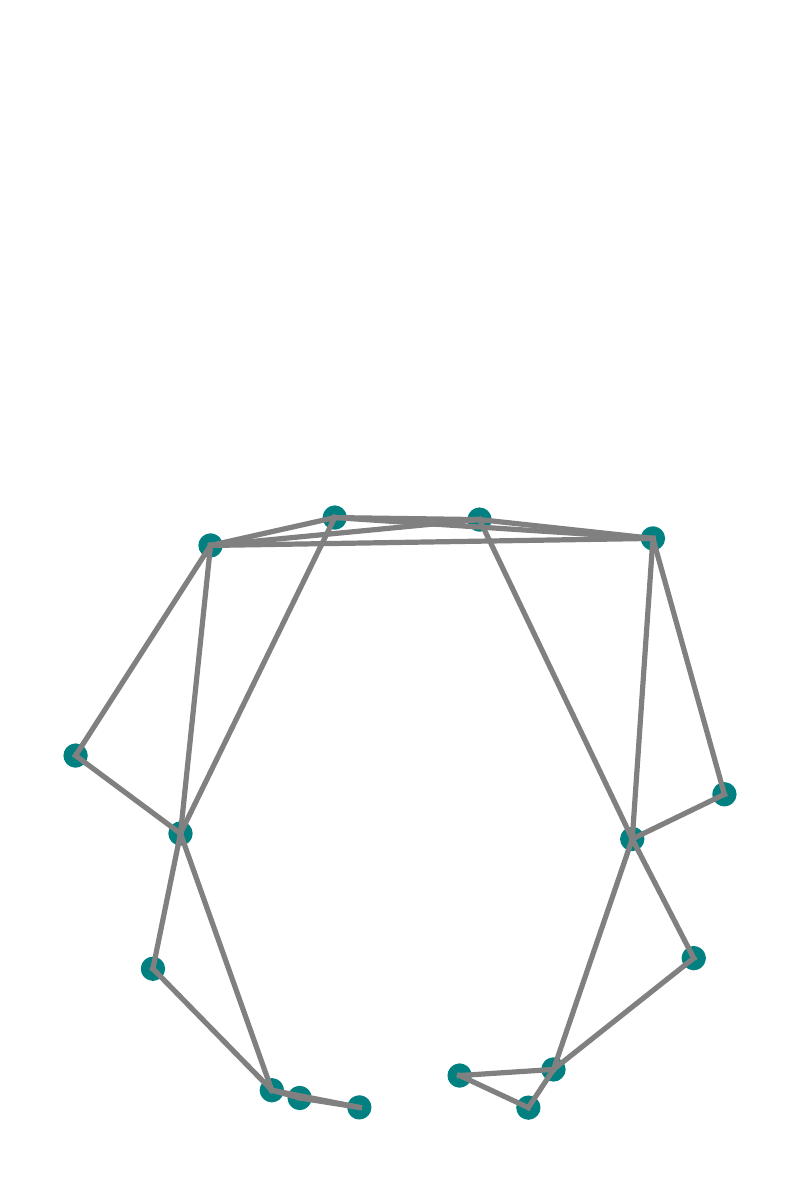}
    \caption{Lower Body}\label{fig:lowerbody}
  \end{subfigure}
  \caption{
	The interaction graphs we used to produce the trajectory datasets.
  }\label{fig:illustration_dataset}
\end{center}
\end{figure}

From these data we produce four different datasets which rely on a different interaction graph sructure. 
Three of these dataset are based on the marker trajectories and one is based on the joint angle trajectories. 
We used three different graph structures which we will respectively call {\em complete skeleton, armless, lower body}.
These graph structures are motivate by the fact that our skeleton provides a nature interation graph. In the complete
skeleton version we track 19 body parts by computing the center of mass of the sensors that are placed on  each body 
part. In the armless version we track 15 body parts; we removed the elbow and hand markers because
these are hard to predict and do not seem to influence the gait dynamics. In the lower body version we use all 
the available markers for the lower body part, i.e. we do not do body part aggregation as in the previous two.
In all three datasets we normalise the trajectories by removing the pelvis position and dividing by the patient height.
The result of this normalization is a patient that seems to walk on a treadmill with position values being in the range \([0,1]\).
Finally in the angle dataset instead of Euclidean trajectories we use the joint angle trajectories of the lower
body resulting in the trajectories of the angles of eight joints. We normalise the angle dataset to the $\mathbcal N(0,1)$.
Note that angles exhibit larger variability than marker position.  We visualise the different structures in 
\ref{fig:illustration_dataset}. 

\subsection{Experimental Setup}
Depending on the dataset, we train cNRI with 128 (or 256) units per 
layer in the unsupervised encoders. The decoder performs the best 
with a hidden size of 256 or 384 units.
This model overfits quickly on the angle dataset due to the small number of samples; we thus reduced 
the number of hidden units to 64 and 128 for the encoder and the decoder respectively.
To avoid overfitting we use a 10\% dropout. We use the Adam optimizer~\cite{Kingma2015} 
with a time-based learning rate decay of $\frac{1}{2}$ every 50 epochs starting at $10^{-3}$.
For the unsupervised encoders, we found that our model generalizes better with two edge-types:
a "non-edge" with a hard-coded prior of 0.91 (the non-edge) and 0.09 for the second edge-type (same as the original NRI).
We evaluate the models using 3-fold cross validation were we divide the dataset to training, validation and test; we 
take care to keep all trajectories of a given patient/visit within one of these sets so that there is no information 
leakage. We tune the hyperparameters on the validation set. 
We report Mean squared error between the real and generated trajectories and its standard deviation 
that we compute over the denormalized generations; the markers' unit is millimeters and the angles' unit is degrees.

We will refer to the various combinations of encoder-decoder of our model as follows: 
{PG-cNRI} is the combination of the perfect interaction graph (PG) encoder with our conditional decoder;
{IG-cNRI} uses the imperfect interaction graph (IG) encoder;
{NRI-cNRI} combines the unsupervised encoder of the NRI with our decoder;
and finally, {fNRI-cNRI} is the combination of fNRI encoder with cNRI.

We compare against several baselines.  The two first baselines are based on the mean. 
Even though simple, they have excellent performance, and on the angles dataset, they 
are hard to beat. The first mean-based baseline predicts for each object its mean on the training set.
The second uses more knowledge and predicts, for each object an object-based average over the side in which the patient is affected.
To avoid errors coming from translation we slide the mean to start at the same position as the trajectory evaluated.
In addition, we use three variants of recurrent neural networks (RNN): standard~\cite{Rumelhart1986}, GRU~\cite{Cho2014} and LSTM~\cite{Hochreiter1997}.
These are autoregressive models that tackle conditional generations heads-on.
We condition their hidden states on the clinical features and train them to
minimize the error between the generated trajectories and the true ones; we 
use no teacher forcing. We also add a reformulation of NRI that can generate
the entire trajectory in which we sample the latent graph from the aggregated
posterior. In our reformulation of the NRI there is no warm-up of the decoder 
state, and the decoder generates directly new states. In addition it uses no
conditioning vector, we included in the experiments in order to verify that
the conditional information does improve generation performance. 
The code associated with this work is available at \url{https://github.com/jacr13/cNRI}.

\subsection{Results}
The models that we propose here are the only ones that consistently beat the improved mean baseline (Table 
\ref{tab:mse}). From the other baselines only the RNN one is able to outperform the improved mean in three of the four 
datasets. On the skeleton dataset, the model that uses the real graph (PG-cNRI) achieves the lowest error.
In PG-cNRI the decoder can only use the skeleton's links to propagate the informationr;
this considerably reduces the model's power for reasoning on long relations.  Since the arms 
are almost unpredictable, PG-cNRI has here the right inductive bias since it propagates  
less information through these nodes making overfitting less likely.
When we remove the arms, (armless, lower body), as expected the performance improves.
Our unsupervised models (NRI-cNRI and fNRI-cNRI) learn better the dynamics and their generations 
are very close to the real trajectories, and they outperform significantly all baselines. 
The angles dataset is the hardest to predict. Here the improved mean is an excellent approximation 
of the real trajectories. Here all our models are better than the improved mean (IG-cNRI being the best), though the 
performance gap is not as large as in the other three dataset.
\begin{table}[ht!]
  \centering
  \setlength{\tabcolsep}{5pt}
  \renewrobustcmd{\bfseries}{\fontseries{b}\selectfont}
  \renewcommand{\pm}{\mathbin{\mbox{\unboldmath$\mathchar"2206$}}}
  \scalebox{0.9}{%
  \begin{tabular}{
    l
    S[detect-weight, table-number-alignment = center, table-format=-3.2(2), mode=text]
    S[detect-weight, table-number-alignment = center, table-format=-3.2(2), mode=text]
    S[detect-weight, table-number-alignment = center, table-format=-3.2(2), mode=text]
    S[detect-weight, table-number-alignment = center, table-format=-3.2(2), mode=text]
  }
  \toprule
  {Model} & {Skeleton} & {Armless} & {Lower Body} & {Angles}\\
  \midrule
  Mean & 477.71 +- 31.63 & 383.09 +- 36.00 & 988.00 +- 258.91 & 85.24 +- 7.79 \\
  Improved Mean & 461.04 +- 25.20 & 356.45 +- 16.68 & 815.58 +- 173.75 & 41.28 +- 3.11 \\
  RNN & 437.82 +- 39.01 & 274.22 +- 15.01 & 776.55 +- 118.68 & 41.60 +- 2.87 \\
  GRU & 527.43 +- 104.98 & 390.50 +- 70.08 & 868.63 +- 119.24 & 41.27 +- 4.71 \\
  LSTM & 556.58 +- 22.28 & 384.49 +- 56.59 & 824.38 +- 94.46 & 41.85 +- 4.02 \\
  NRI & 538.71 +- 5.95 & 354.67 +- 21.34 & 827.81 +- 74.12 & 41.42 +- 3.62 \\
  \midrule
PG-cNRI & \bfseries 380.42 +- 43.71 & 302.68 +- 64.68 & 772.75 +- 113.13 & 38.19 +- 2.50 \\
IG-cNRI & 474.27 +- 122.62 & 351.33 +- 116.93 & 856.64 +- 150.88 & \bfseries 37.89 +- 2.30 \\
  NRI-cNRI & 399.83 +- 67.07 & \bfseries 212.97 +- 20.21 & \bfseries 696.09 +- 113.42 & 39.60 +- 2.35 \\
  fNRI-cNRI & 433.87 +- 133.00 & 241.93 +- 19.23 & \bfseries 696.47 +- 58.93 & 40.83 +- 3.11 \\    
  \bottomrule
  \end{tabular}%
  }
  \caption{MSE and std of conditional generations.}
  \label{tab:mse}
\end{table}

\begin{figure}
  \centering
  \begin{subfigure}[b]{0.3\textwidth} 
      \centering \includegraphics[width=\textwidth]{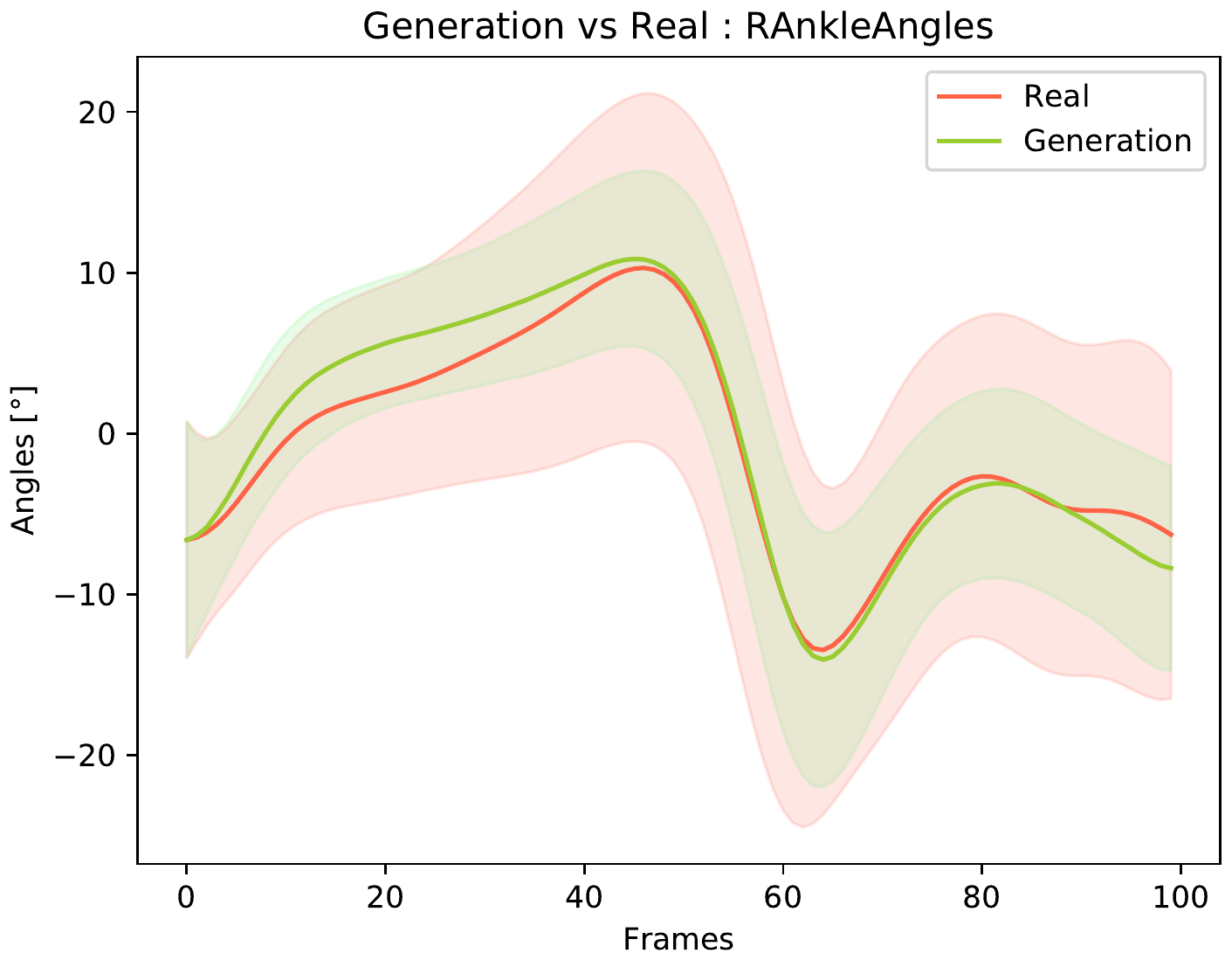}
      \caption{Right Ankle}\label{fig:ankle}
  \end{subfigure}
  ~
  \begin{subfigure}[b]{0.3\textwidth}
      \centering \includegraphics[width=\textwidth]{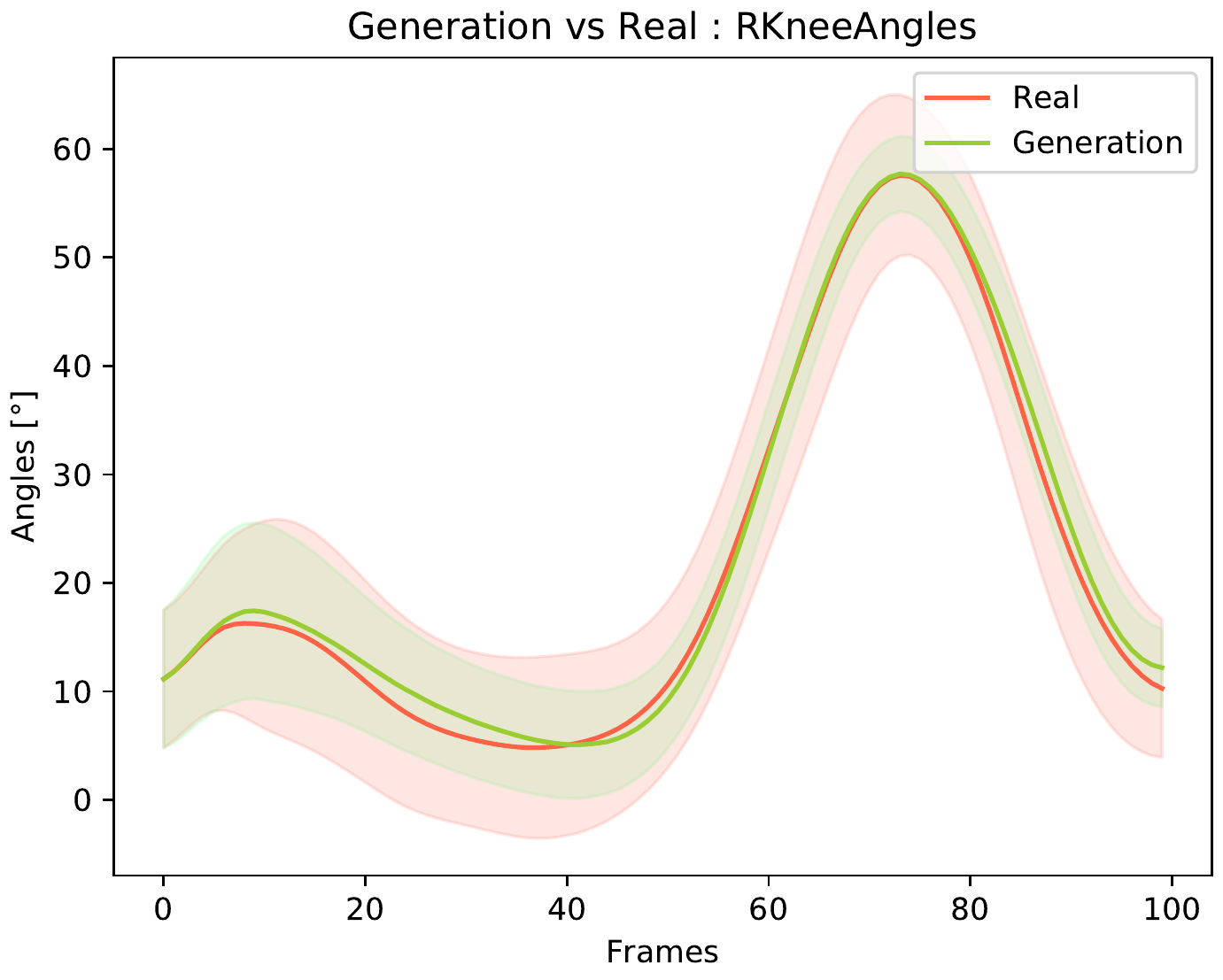}
      \caption{Right Knee}\label{fig:knee}
  \end{subfigure}
  ~
  \begin{subfigure}[b]{0.3\textwidth}
      \centering \includegraphics[width=\textwidth]{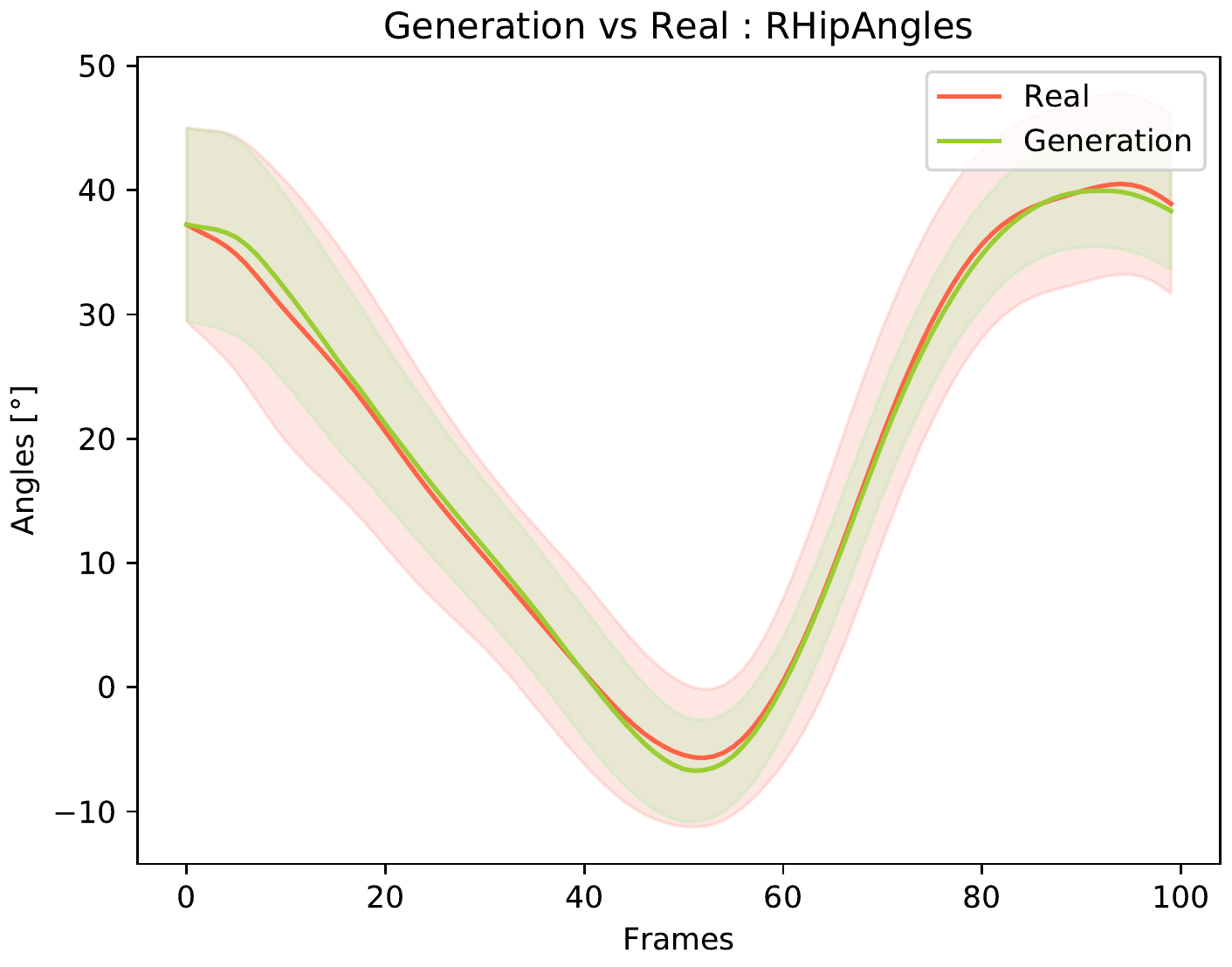}
      \caption{Right Hip}\label{fig:hip}
  \end{subfigure}
\caption{Mean and variance of right side angle generations with IG-cNRI model.}\label{fig:angles_right}
\end{figure}

\begin{figure}
  \begin{subfigure}{\linewidth}
    \centering
    \includegraphics[scale=0.18]{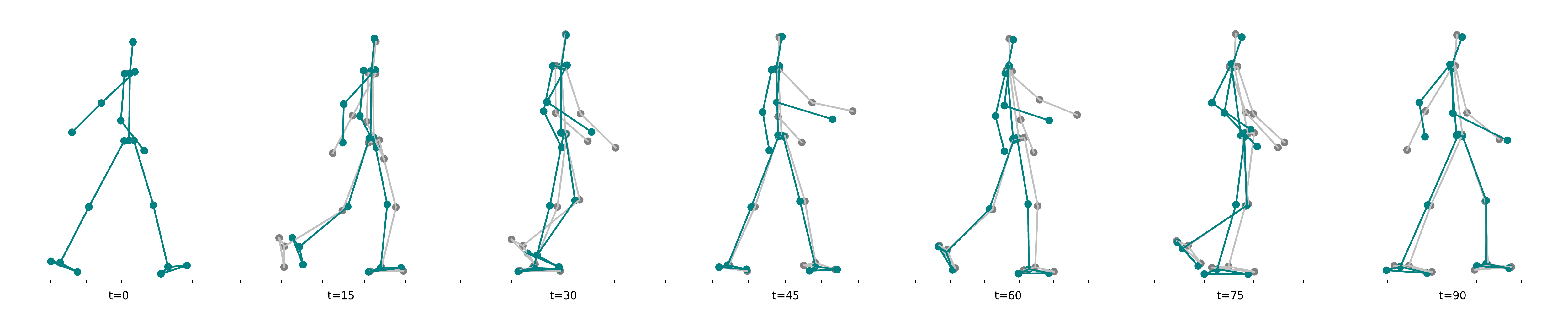}
    \includegraphics[scale=0.18]{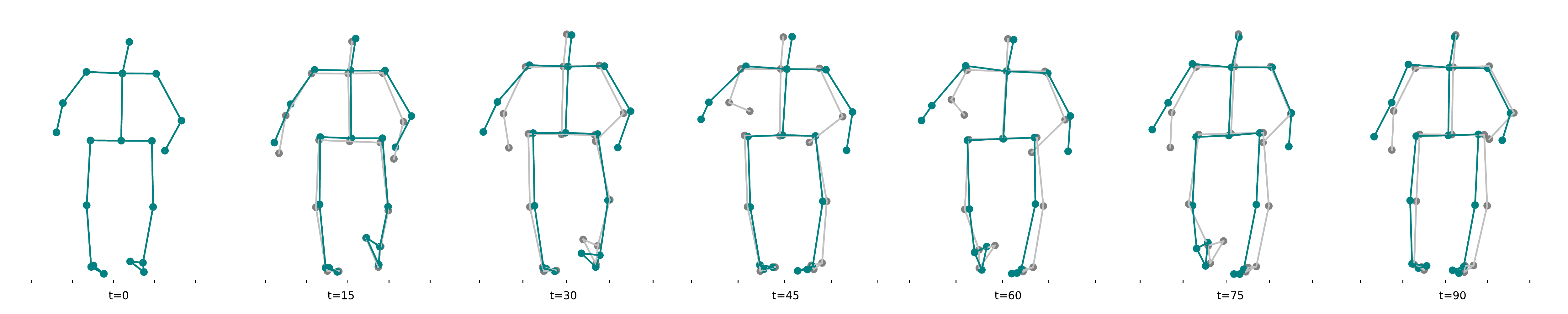}
    \caption{Example of PG-cNRI generations on skeleton dataset.}
    \label{fig:gen_all}
  \end{subfigure}
  \begin{subfigure}{\linewidth}
    \centering
    \includegraphics[scale=0.18]{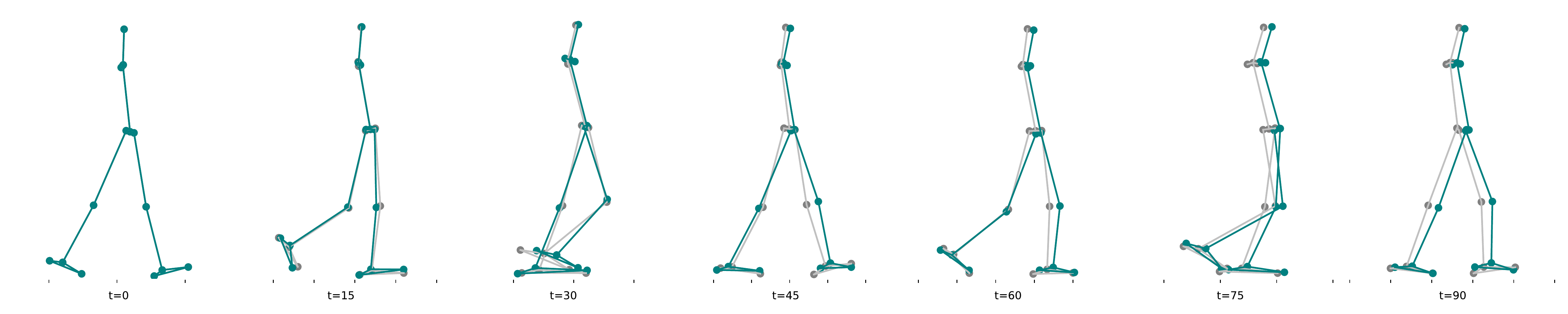}
    \includegraphics[scale=0.18]{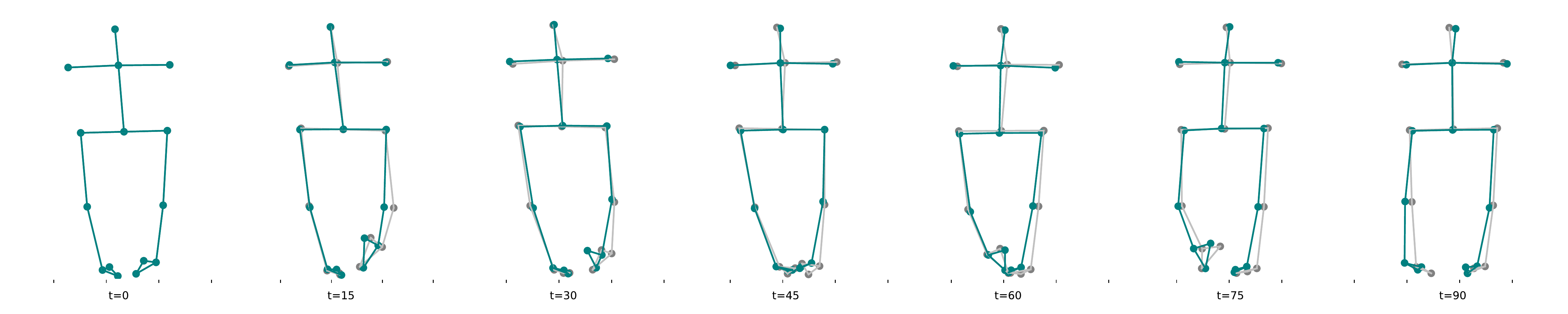}
    \caption{Example of NRI-cNRI generations on armless dataset.}
    \label{fig:gen_armles}
  \end{subfigure}
  \begin{subfigure}{\textwidth}
    \centering
    \includegraphics[scale=0.18]{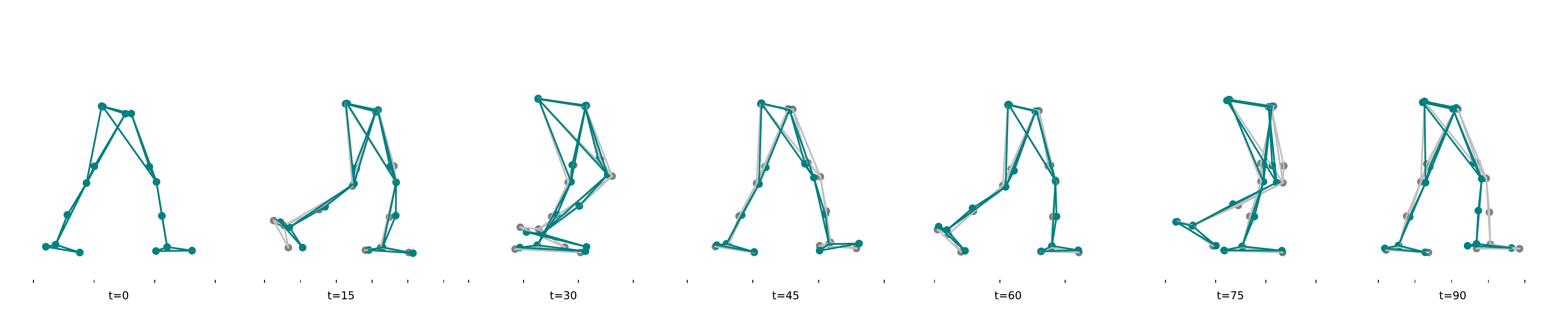}
    \includegraphics[scale=0.18]{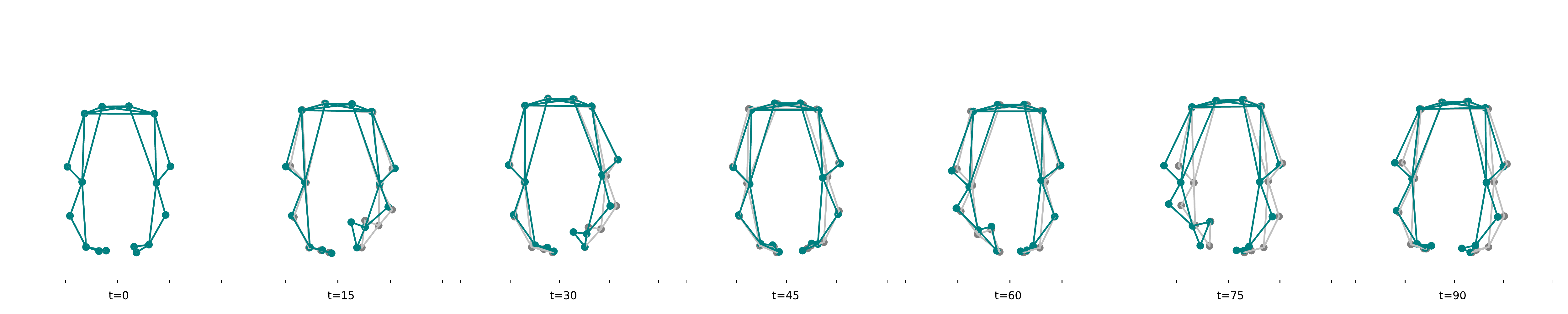}
    \caption{Example of NRI-cNRI generations on lower body dataset.}
    \label{fig:gen_lb}
  \end{subfigure}
  \caption{Generations (in blue) against real trajectories (in gray), the edges are from the real graph, not \(z\).}
  \label{fig:gen}
\end{figure}

We give examples of generations in  Figure \ref{fig:angles_right} and Figure \ref{fig:gen}, where we see that these are very close to the real ones.
In Figure \ref{fig:angles_right} we report the angle trajectories, mean and standard deviation over the test set, for the real and generated data 
for the angles located in the right side of the body. Our model follows nicely the dynamics, but in some cases its trajectories have less variance 
than the real one.  In Figure \ref{fig:gen} we provide snapshots of body positions for the real and generated data. 

We notice that some of the baselines and cNRI models have high error variance. This is the result of the 
variable number of gait cycles we have per patient and the fact that we have patients that are affected
on different body sides, left or right.  When we split the data for evaluation we take care
that a patient's  are only present in one of the training, validation, test sets. 
As a result splits can be unbalanced with respect to the affected body side, which increases the 
risk of overfitting, with the underrepresented side in the training set leading to poor generations in the testing phase.  
This is someting that we indeed verified by looking at the errors and distributions of the affected sides over the folds.

\begin{figure}
  \centering
  \begin{subfigure}[b]{0.3\textwidth} 
      \centering \includegraphics[width=\textwidth]{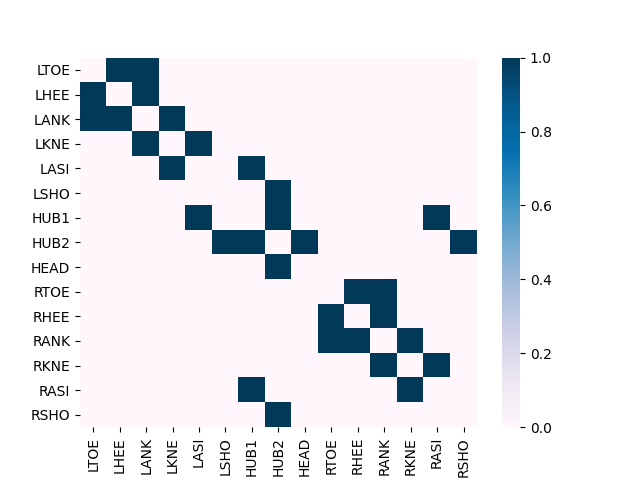}
      \caption{PG-cNRI}\label{fig:perfectGraph}
  \end{subfigure}
  \begin{subfigure}[b]{0.3\textwidth}
      \centering \includegraphics[width=\textwidth]{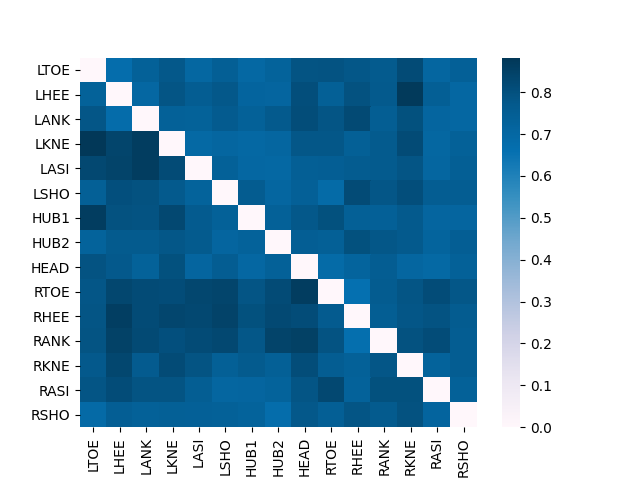}
      \caption{IG-cNRI}\label{fig:imperfectGraph}
  \end{subfigure}
  \begin{subfigure}[b]{0.3\textwidth}
      \centering \includegraphics[width=\textwidth]{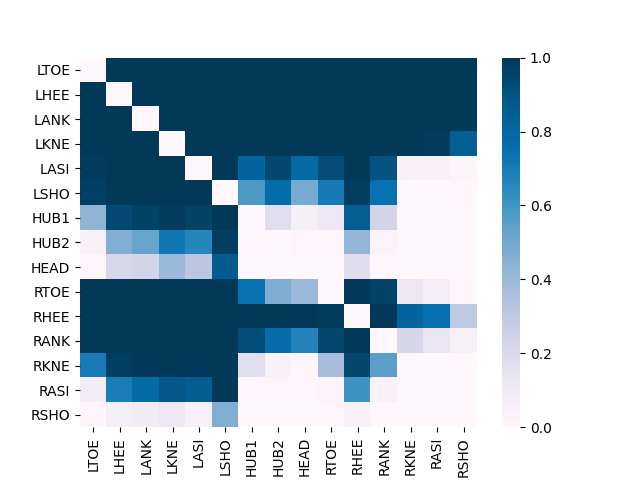}
      \caption{NRI-cNRI}\label{fig:hip}
  \end{subfigure}
  \centering
  \begin{subfigure}[b]{0.24\textwidth} 
      \centering \includegraphics[width=\textwidth]{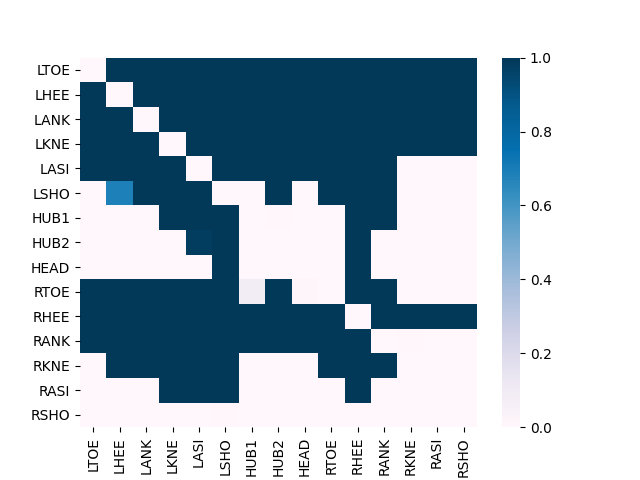}
      \caption{left}\label{fig:0-left}
  \end{subfigure}
  \begin{subfigure}[b]{0.24\textwidth}
      \centering \includegraphics[width=\textwidth]{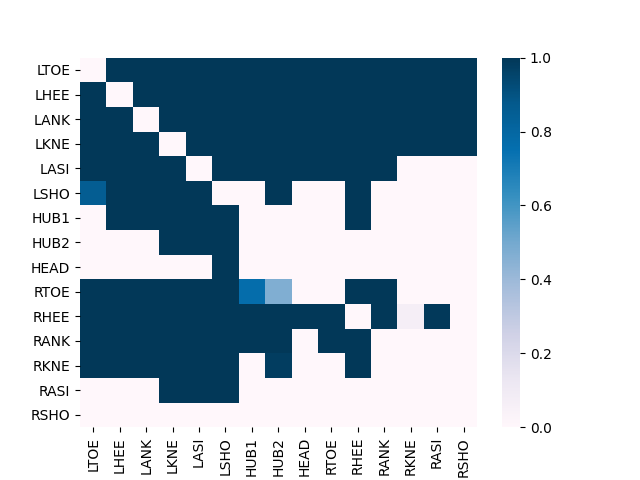}
      \caption{left}\label{fig:2-left}
  \end{subfigure}
  \begin{subfigure}[b]{0.24\textwidth}
      \centering \includegraphics[width=\textwidth]{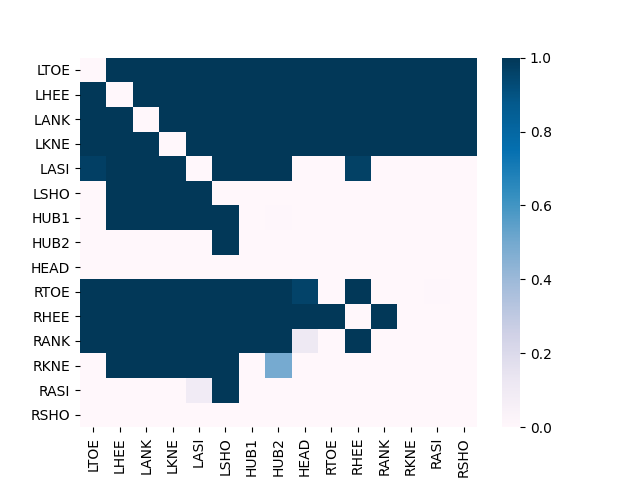}
      \caption{right}\label{fig:1-right}
  \end{subfigure}
  \begin{subfigure}[b]{0.24\textwidth}
      \centering \includegraphics[width=\textwidth]{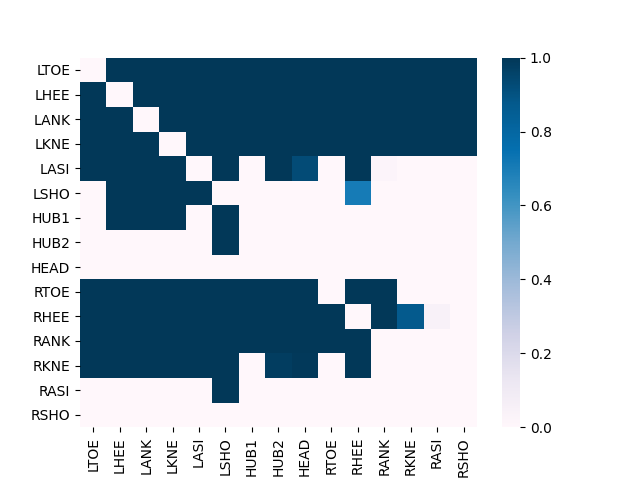}
      \caption{right}\label{fig:3-right}
  \end{subfigure}
\caption{Interaction maps. Top row interaction maps used in the three methods; for NRI-cNRI we give the aggregate posterior.
We do not inlude fNRI-cNRI for space reasons due to the large number of edge types. Bottom row: patient specific interaction 
graphs for NRI-cNRI, first row two random patients with left hemiplegia, second row 
right hemiplegia.}\label{fig:interaction_graphs}
\end{figure}

Finally in the top row of Figure~\ref{fig:interaction_graphs} we give the interaction maps that are used in PG-cNRI (given), 
IG-NRI (learned) and NRI-cNRI (aggregate posterior); we do not include fNRI-cNRI because it has one adjacency 
matrix per edge type and lack of space. As we see IG-NRI establishes a non-sparse interaction matrix where 
every part interacts with every other part. In NRI-cNRI the picture that arises from the aggregate posterior
is much sparser. In the bottom row of the same figure we give the interaction maps established by NRI-cNRI 
for particular patients (random patients with left and right hemiplegia). We see that even though they 
are all quite close to the aggregate posterior structure there exist systematic structural differences between 
patients  with left and right hemiplegia. This points to future improvements of the model where we can introduce
dependency structures between the condition vector and the interaction maps, using hiererachical models which we
will allow as to have more informed priors that are conditioned on $\mathbf c$.

\section{Conclusion}
Motivated by the need for decision support in the treatment of patients with gait
impairements we propose a conditional generative model, based on an extenstion NRI ~\cite{Kipf2018},
that can learn to conditionally generate from different physical systems.
Our model has two components: the first is an encoder that learns a graph of
interactions of the different body parts.
The second is a decoder that uses the interaction graph together with a conditioning  
vector that describes the specificities of the particular dynamical system and learns 
the conditional dynamics.  The experiments show that the proposed model outperforms 
the baselines in all dataset we experimented.
Moreover the method achieves very good performance even though it has been trained on 
relatively small training datasets, in fact very small when it comes to the typical 
training size used in deep learning generative models. This is an important feature
of the method since many applications, such as the one we explored here, the available
training data will be very limited. As a future work we want to explore different 
structures in the inference and generative models and different dependence assumptions
in order to increase further the generation quality, e.g. diferent dependency structures
between the interaction matrix and the conditioning vector and/or learning to predict
the latter from the former. 

\section*{Acknowledgments}
This work was supported by the Swiss National  Science	Foundation  grant
number CSSII5\_177179 ``Modeling pathological gait resulting from motor impairment''.

%
%
%
\bibliographystyle{splncs04}

\end{document}